\documentclass[letterpaper, 10 pt, conference]{ieeeconf}  % Comment this line out if you need a4paper

\IEEEoverridecommandlockouts                              % This command is only needed if 
\usepackage{graphics} % for pdf, bitmapped graphics files
\usepackage{epsfig} % for postscript graphics files
\usepackage[utf8]{inputenc} % allow utf-8 input
\usepackage[T1]{fontenc}    % use 8-bit T1 fonts
\usepackage{hyperref}       % hyperlinks
\usepackage{url}            % simple URL typesetting
\usepackage{amsmath,amssymb,amsfonts}
\usepackage{nicefrac}       % compact symbols for 1/2, etc.
\usepackage{microtype}      % microtypography
\usepackage{physics}
\usepackage{algorithm, algpseudocode}
\usepackage{paralist}
\usepackage[disable]{todonotes}

\title{\LARGE \bf
Automatic Differentiation and Continuous Sensitivity Analysis\\
of Rigid Body Dynamics
}

\author{David Millard${}^*{}^1$\thanks{${}^*$ Equal contribution}\!\!, Eric Heiden${}^*{}^1$\!\!, Shubham Agrawal${}^2$\!\!,
Gaurav S. Sukhatme${}^1$% <-this % stops a space
\thanks{$^{1}$David Millard, Eric Heiden, Gaurav S. Sukhatme are with the Department of Computer Science, University of Southern California, Los Angeles, USA
        {\tt\small \{dmillard,heiden,gaurav\}@usc.edu}}%
\thanks{$^{2}$ Shubham Agrawal is with the Department of Computer Science, Columbia University, New York, USA
        {\tt\small sa3762@columbia.edu}
        }%
}

\usepackage{xparse}

\newcommand{\Q}{\mathbf{q}}
\newcommand{\Qd}{\dot{\mathbf{q}}}
\newcommand{\Qdd}{\ddot{\mathbf{q}}}
\NewDocumentCommand\transform{mg}{\ensuremath{%
    \IfNoValueTF{#2}{%
        \boldsymbol{X}_{#1}%
    }{%
        {}^{#2}\boldsymbol{X}_{#1}%
    }%
}}
\NewDocumentCommand\transformJ{mg}{%
    \IfNoValueTF{#1}{%
        \transform{\text{J}}%
    }{%
        \transform{\text{J}#1}%
    }%
}

\newcommand{\jsim}{\mathbf{H}}
\newcommand{\forces}{\ensuremath{\boldsymbol{\tau}}}
\newcommand{\params}{\ensuremath{\boldsymbol{\theta}}}
\newcommand{\state}{\ensuremath{\mathbf{x}}}
\newcommand{\control}{\ensuremath{\mathbf{u}}}
\newcommand{\dstate}{\ensuremath{\dot{\mathbf{x}}}}

\newcommand{\StateSpace}{\ensuremath{\boldsymbol{\mathfrak{X}}}}
\newcommand{\ControlSpace}{\ensuremath{\boldsymbol{\mathfrak{U}}}}
\newcommand{\ParamSpace}{\ensuremath{\boldsymbol{{\Theta}}}}
\newcommand{\Loss}{\mathcal{L}}

\DeclareMathOperator*{\argmin}{arg\,min}

\begin{document}

\maketitle
\thispagestyle{empty}
\pagestyle{empty}

%%%%%%%%%%%%%%%%%%%%%%%%%%%%%%%%%%%%%%%%%%%%%%%%%%%%%%%%%%%%%%%%%%%%%%%%%%%%%%%%
\begin{abstract}
A key ingredient to achieving intelligent behavior is physical understanding that equips robots with the ability to reason about the effects of their actions in a dynamic environment. Several methods have been proposed to learn dynamics models from data that inform model-based control algorithms. While such learning-based approaches can model locally observed behaviors, they fail to generalize to more complex dynamics and under long time horizons.

%Under the umbrella of intuitive physics, specialized models, such as interaction and graph neural networks, have been proposed to learn dynamics from data to predict the motion of objects over long time horizons.  By labelling the training data given to these models by physical quantities, they are able to produce behavior that is conditioned on actual physical parameters, such as masses or friction coefficients, allowing for plausible estimation of physical properties and improved generalizability.

In this work, we introduce a differentiable physics simulator for rigid body dynamics. Leveraging various techniques for differential equation integration and gradient calculation, we compare different methods for parameter estimation that allow us to infer the simulation parameters that are relevant to estimation and control of physical systems. In the context of trajectory optimization, we introduce a closed-loop model-predictive control algorithm that infers the simulation parameters through experience while achieving cost-minimizing performance.

% A differentiable physics engine provides many advantages when used as part of a learning process. Physically accurate simulation obeys dynamics laws of real systems, including conservation of energy and momentum. Furthermore, joint constraints are enforced with no room outside of the model for error. The parameters of a physics engine are well-defined and correspond to properties of real systems, including multi-body geometries, masses, and inertia matrices. Learning these parameters provides a significantly interpretable parameter space, and can benefit classical control and estimation algorithms. Further, due to the high inductive bias, model parameters need not be jointly retrained for differing degrees of freedom or a reconfigured dynamics environment.
\end{abstract}
%%%%%%%%%%%%%%%%%%%%%%%%%%%%%%%%%%%%%%%%%%%%%%%%%%%%%%%%%%%%%%%%%%%%%%%%%%%%%%%%

\section{INTRODUCTION}

Physically-based reasoning is fundamental to successfully performing complex tasks in the physical world. This is particularly relevant to the domain of robot learning. There is a large body of mature work in robot dynamics, which need not be learned from scratch per task. In this work, we introduce a differentiable physics simulator for rigid body dynamics. We leverage this differentiability to estimate parameters that result in simulations that closely match the behavior of observed reference systems. Additionally, through trajectory optimization, we can efficiently generate control inputs that minimize cost functions that are expressed with respect to any quantity that is part of the physics computation.

Differentiable physics provides many advantages when used as part of a learning process. Physically accurate simulation obeys dynamical laws of real systems, including conservation of energy and momentum. Furthermore, joint constraints are enforced with no room for error. The parameters of physics engines, like link geometry and inertia matrices, are well-defined and correspond to properties of real systems. Learning these parameters provides a significantly interpretable parameter space, and can benefit classical control and estimation algorithms. These systems provide a high inductive bias, and model parameters need not be retrained for different tasks or reconfigured environments.

Our contributions are as follows:
\begin{compactenum}
    \item We present a fully differentiable simulator for rigid body dynamics that supports a rich set of integrators for the accurate simulation of mechanisms over long time horizons.
    \item We analyze the performance of gradient calculation methods on the problem of inferring parameters underlying the simulation of rigid-body dynamical systems.
    \item We introduce an adaptive model-predictive control algorithm that leverages our differentiable model to perform trajectory optimization while finding the optimal parameters that fit a reference mechanism implemented in another simulator.
\end{compactenum}

% For exact calculation of derivative information, we use techniques from the automatic differentiation (AD) and sensitivity analysis literature. We present several ways of computing gradients through rigid-body dynamics equations, and compare the computational tradeoffs of each approach.

\begin{figure}[t]
    \centering
    \newcommand{\teaserwidth}{.4\columnwidth}
    \newcommand{\margin}{\hspace{.5cm}}
    \margin
    \includegraphics[trim=12cm 0cm 12cm 0cm,clip,width=\teaserwidth]{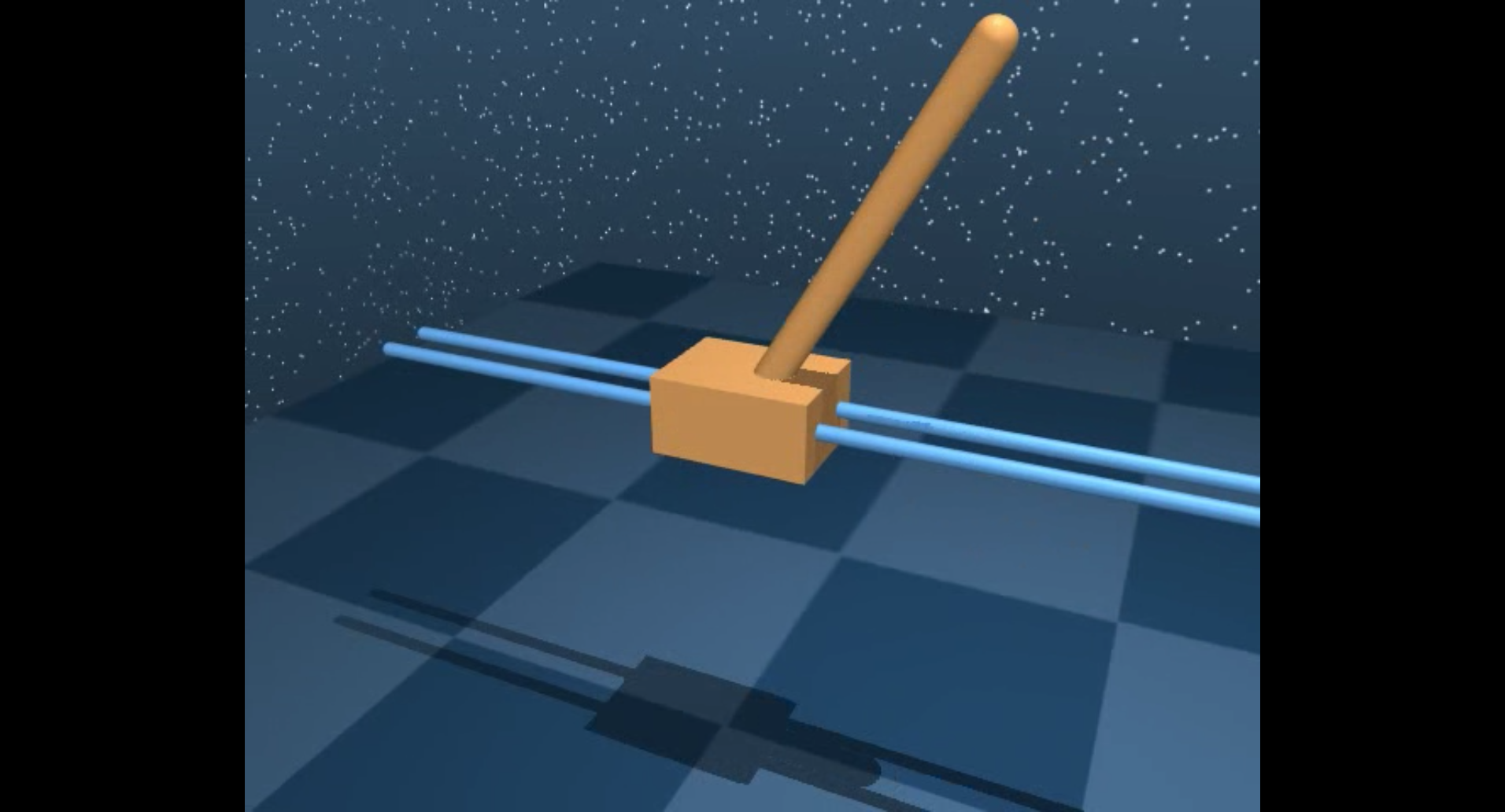}\hfill
    \includegraphics[trim=15cm 2cm 15cm 0cm,clip,width=\teaserwidth]{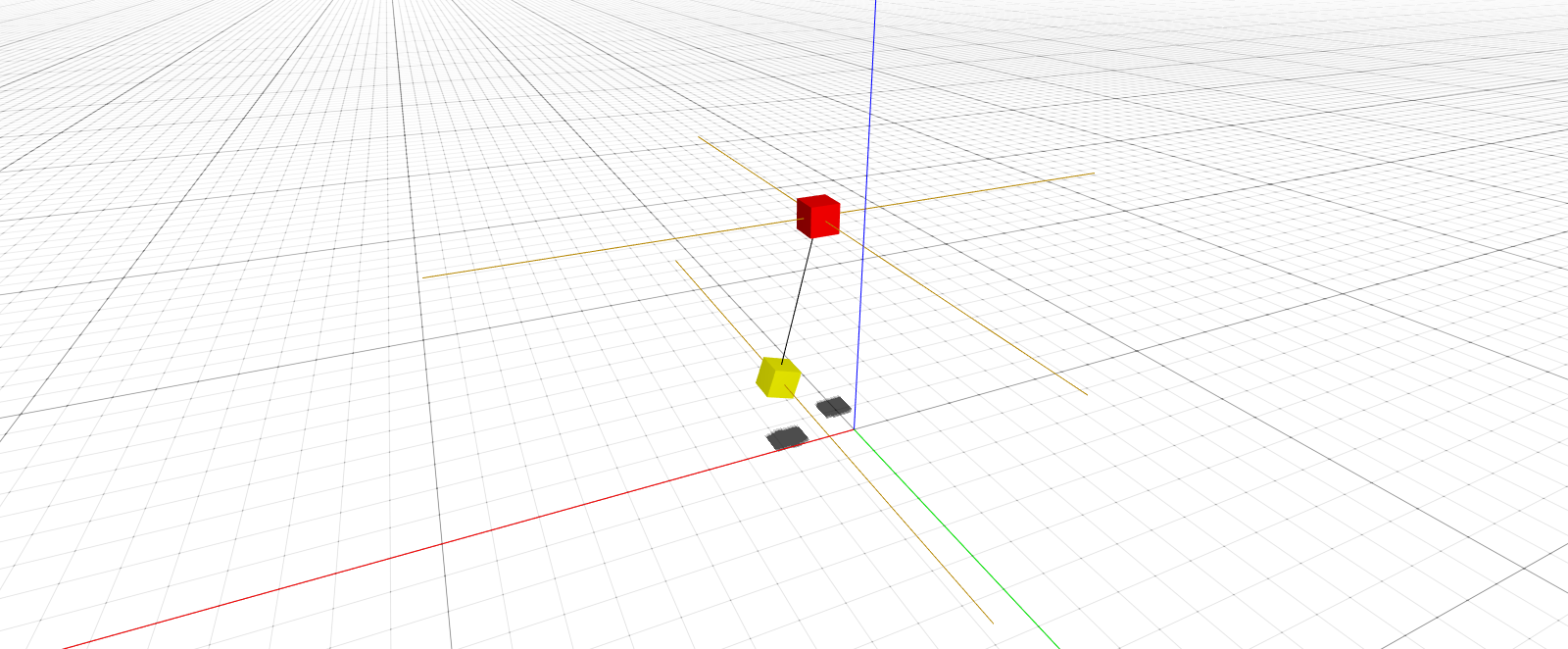}
    \margin~\\
    \margin
    \includegraphics[trim=4cm 3cm 4cm 7cm,clip,width=\teaserwidth]{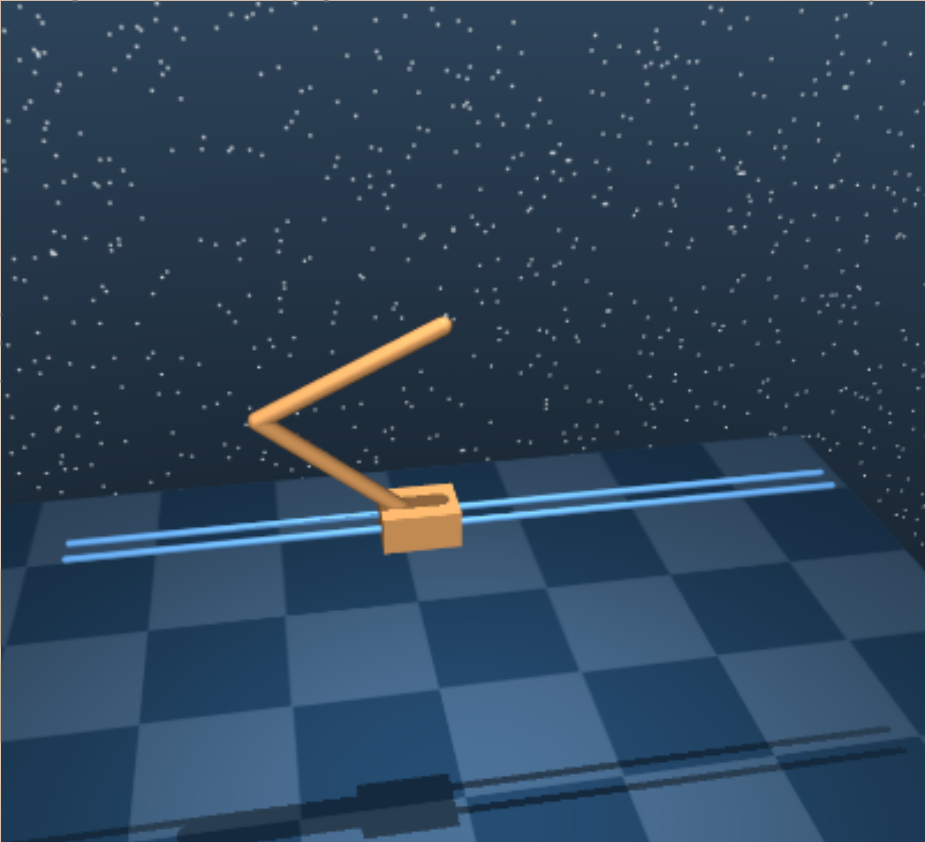}\hfill
    \includegraphics[trim=15cm 2cm 15cm 4cm,clip,width=\teaserwidth]{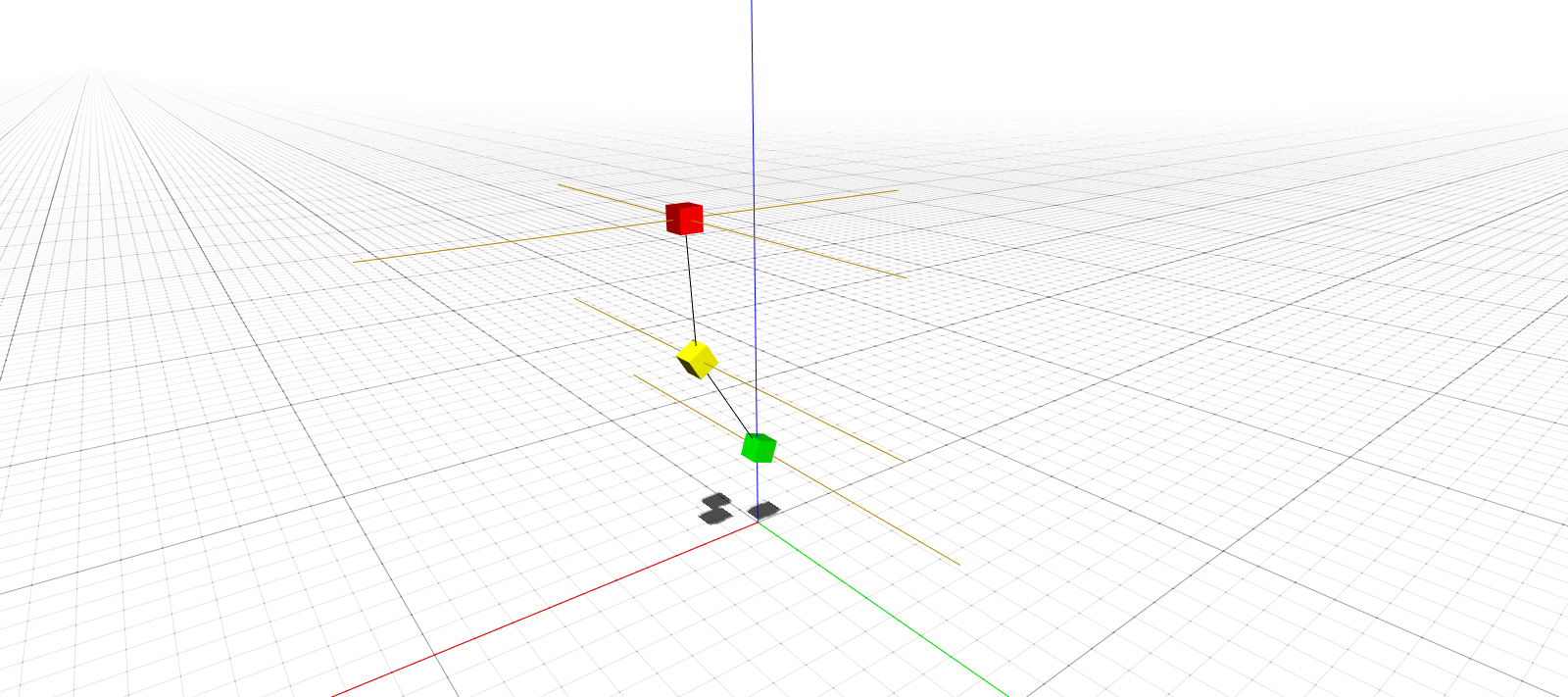}
    \margin
    \caption{Visualizations of the simulated cartpole systems. The carts are constrained to the rail, but may move infinitely in either direction. Both systems are actuated by a linear force applied to the cart.
    \textit{Upper row:} Single cartpole environment from DeepMind Control Suite~\cite{tassa2018dm} in the MuJoCo~\cite{todorov2012mujoco} physics simulator (left) and in our simulator (right). 
    \textit{Lower row:} Double cartpole environment from DeepMind Control Suite (left) and in our simulator (right).}
    \label{fig:models}
\end{figure}

\section{RELATED WORK}
\label{sec:related}
Differentiable physics has recently attracted significant research efforts. Degrave et al.~\cite{degrave2019physics} implemented a differentiable physics engine in the automatic differentiation framework Theano. Giftthaler et al.~\cite{giftthaler2017autodiff} presented a rigid-body-dynamics simulator that allows for the computation of derivatives through code generation via RobCoGen~\cite{frigerio2016robcogen}. Similarly, we use Stan Math~\cite{carpenter2015stan}, a C++ library for reverse-mode automatic differentiation to efficiently compute gradients, even in cases where the code branches significantly. Analytical gradients of rigid-body dynamics algorithms have been implemented in the Pinnocchio library~\cite{carpentier2018analytical} to facilitate optimal control and inverse kinematics. While such manually derived gradients can be computed efficiently, they are less general than our approach since they can only be used to optimize for a number of hand-engineered quantities. More recently, Koolen and Deits~\cite{koolen2019rbd-julia} have implemented rigid-body-dynamics algorithms in the programming language Julia where, among others, libraries for optimization, automatic differentiation, and numerical integration are available. Non-penetrative multi-point contacts between rigid bodies are often simulated by solving a linear complementarity problem (LCP), through which~\cite{peres2018lcp} differentiate using the differentiable quadratic program solver OptNet~\cite{amos2017optnet}. While our proposed model does not yet incorporate contact dynamics, we are able to demonstrate the scalability of our approach on versatile applications of differentiable physics to common 3D control domains.

Automatic differentiation of the solutions to differential equations is well studied, with applications to pharmacology, meteorology, and many other fields. Recent machine learning work by ~\cite{chen2018neural} recasts learning in long short-term memory networks and residual networks as approximations to this problem. Thorough comparisons of methods for computing parameter gradients of ODE solutions are given in ~\cite{carpenter2015stan, rackauckas2018comparison, serban2003cvodes}.

Learning dynamics models has a tradition in the field of robotics and control theory. Early works on forward models~\cite{moore1992forward} and locally weighted regression~\cite{atkeson1997lwl} yielded control algorithms that learn from previous experience. Computing gradients through the solution of differential equations has been further leveraged for system identification~\cite{rackauckas2018comparison}.

More recently, a variety of novel deep learning architectures have been proposed to learn \emph{intuitive physics} models. Inductive bias has been introduced through graph neural networks~\cite{hadsell2018graphnet, li2018learning, liu2019psd}, particularly interaction networks~\cite{battaglia2016interaction, chang2016physics, schenck2018spnet, mrowca2018flexible, xu2019physics} that are able to learn rigid and soft body dynamics. Vision-based machine learning approaches to predict the future outcomes of the state of the world have been proposed~\cite{wu2015galileo, wu2016phys101, wu2017deanimation, finn2016unsupervised, janner2018reasoning}. \emph{Physics-informed learning} imposes a stronger inductive bias on the learning problem to model particular physical processes, such as cosmological structure formation~\cite{he2019physics} or turbulence models~\cite{raissi2018physics}. Deep Lagrangian Networks~\cite{lutter2018delan} and Hamiltonian Networks~\cite{greydanus2019hamiltonian} represent functions in the respective classical mechanics frameworks using deep neural networks.

The approach of adapting the simulator to real world dynamics, which we demonstrate through our adaptive MPC algorithm in \autoref{sec:ampc}, has been less explored. While many previous works have shown to adapt simulators to the real world using system identification and state estimation~\cite{kolev2015sysid, zhu2018fastmi}, few have shown adaptive model-based control schemes that actively close the feedback loop between the real and the simulated system~\cite{reichenbach2009dynsim, farchy2013simback, chebotar2018sim2real}. Instead of using a simulator, model-based reinforcement learning is a broader field~\cite{polydoros2017mbrl}, where the system dynamics, and state-action transitions in particular, are learned to achieve higher sample efficiency compared to model-free methods. Within this framework, predominantly Gaussian Processes~\cite{ko2007gp, deisenroth2011pilco, boedecker2014sgp} and neural networks~\cite{williams2016mppi, yamaguchi2016neural} have been proposed to learn dynamics and optimize policies. Bayesian neural networks in particular have been used to learn dynamics in model-based reinforcement learning approaches~\cite{fu2016one, depeweg2016mbrl, gal2016improving, chua2018deep}.

\section{NOTATION}
\label{sec:notation}
\noindent
Throughout this work, we follow the following conventions.
$\state\in\StateSpace$ denotes the system's state vector from the state space $\StateSpace$.
$\control\in\ControlSpace$ denotes the system's control vector from the control space $\ControlSpace$.
$\params\in\ParamSpace$ denotes the system's parameter vector from the parameter space $\ParamSpace$.
A rigid body system is entirely described by the generalized coordinates\footnote{Generalized coordinates sparsely encode only particular degrees of freedom in the kinematic chain so that connected bodies remain connected.} $\forces, \Q, \Qd, \Qdd$, which denote the generalized forces, positions, velocities, and accelerations, respectively.
\section{APPROACH}
\label{sec:approach}

\subsection{Rigid Body Dynamics}

To simulate the dynamics of a rigid body system, we integrate the Newton-Euler equation
\begin{equation}
\label{eqn:newton-euler}
\forces = \jsim(\Q)\Qdd + C(\Q, \Qd) + G(\Q).
\end{equation}
$\jsim(\Q)$ gives the generalized inertial matrix of the system for configuration $\Q$, and $C(\Q, \Qd)$ describes the centrifugal and Coriolis terms affecting motion. $G(\Q)$ describes the contribution from gravity.

% \begin{figure}
%     \centering
%     \includegraphics[width=\columnwidth]{fig/pytorch_autograd_function.pdf}
%     \caption{IDS deep learning layer with its possible inputs and outputs, unrolling our proposed dynamics model over $H$ time steps to compute future system quantities given current joint coordinates, joint forces $\mathbf{\tau}_t$ and model parameters $\theta$. $\operatorname{FD}(\cdot)$ computes the joint velocities, $\operatorname{INT}(\cdot)$ integrates accelerations and velocities, and $\operatorname{KIN}(\cdot)$ computes the 3D positions of all objects in the system in world coordinates. Depending on which quantities are of interest, only parts of the inputs and outputs are used. In some use cases, the joint forces are set to zero after the first computation to prevent their repeated application to the system. Being entirely differentiable, gradients of the motion of objects w.r.t the input parameters and forces are available to drive inference, design and control algorithms.}%     \label{fig:pytorch_autograd_function}
% \end{figure}

Given a descriptive model consisting of joints, bodies, and predecessor/successor relationships, we build a kinematic chain that specifies the dynamics of the system.
From a mechanism's position vector $\Q$, the forward kinematics function $\operatorname{KIN}(\cdot)$ computes the positions and orientation quaternions of the geometries attached to the kinematic chain (such as the end-effector of a robot arm) in world coordinates.

Forward dynamics, computed by $\operatorname{FD}(\cdot)$ is the mapping from positions, velocities and forces to accelerations. We efficiently compute forward dynamics using the Articulated Body Algorithm (ABA) \cite{featherstone2007rbda}, that propagates forces through the bodies while adhering to the motion subspaces defined by the joints that connect them.
In our simulator, bodies comprise physical entities with mass, inertia, and attached rendering and collision geometries. Joints describe constraints on the relative motion of bodies in a model.  Equipped with such a graph of $n$ bodies connected via joints with forces acting on them, ABA computes the joint accelerations $\mathbf{\ddot{q}}$ in $O(n)$ operations.

\subsection{Integration}

Unless specified otherwise, we represent the mechanism's state $\state$ by $[\Q, \Qd]$ so that the change in state $\dstate$ corresponds to $[\Qd, \Qdd]$. A mechanism is parameterized by the real vector $\params$. Such parameters can, depending on the particular system, contain values for the geometries of the links, inertia properties, and other settings that influence the dynamics of the mechanism.
% The control vector $\control$ is typically characterized by the applied joint forces $\forces$.

Resulting from the forward dynamics $f$, a new change in state $\dstate(t_i)$ is computed using ABA at each time step $t_i$ given the previous state $\state(t_{i-1})$ and parameters $\params$. Such relationship forms an ordinary differential equation (ODE)
$
\dstate(t_i) = f(\state(t_{i-1}), t_i, \params)
$, which is solved for the next state $\state(t_i)$ through an integrator. We leverage several methods to solve ODEs, from a simple Euler integrator, through explicit stepping schemes like fourth-order Runge-Kutta (RK4), to adaptive stepping algorithms, such as Dormand-Prince (commonly referred to as RK45) method.

In order to simulate a system, the ODE is solved for a sequence of time steps $t_i\in[t_0,\dots,t_T]$. Throughout this work we consider equidistant time intervals with an integration step size $\Delta t$. The smaller the step size, the more accurate the simulation, but the more ODE system evaluations are necessary. Larger time steps improve the execution performance of the simulator but yield decreased accuracy, particularly in chaotic systems.

\subsection{Parameter Estimation}
\label{sec:param_estimation}
\todo{complete}

We are interested in the behavior of this ODE system with respect to changes in its parameters, $\params$, and to its control inputs, $\control$. Parameters of note are the continuous parameters describing the inertia and geometry of links and joint attachments. Discrete parameters, describing the structure of the system, are fundamentally interesting, but are not considered in this method.

Given a cost function $c: \StateSpace\to\mathbb{R}$ evaluated over states evaluated at times $t_i$, the overall loss $\Loss$ is defined as follows:
\begin{align}
\label{eq:adj_loss}
    \Loss = \sum_t c(\state(t_i)).
\end{align}
Note that the gradient $\pdv{\Loss}{\params}$ only becomes available by integrating over the dynamics $f$ so that the parameters influence the system states. Typically, for parameter estimation, the loss is the distance $||\state(t_i)-\state^*(t_i)||_2^2$ between the simulated states $\state(t_i)$ and the states from a reference trajectory $\state^*(t_i)$, which can be given from a real physical system or another simulation with unknown parameters.
This approach is known as an \emph{initial value problem} (IVP) and has an important application in simulation-to-reality (sim2real) transfer learning, where the reality gap between the agent's dynamics model and the real world dynamics needs to be sufficiently small for the agent to operate in the real world (cf.~\cite{polydoros2017mbrl}).

\subsection{Analytical Differentiation}
In order to estimate parameters, we seek to minimize \autoref{eq:adj_loss} through gradient-based optimization. Such an approach requires calculating gradients of the parameters $\params$ with respect to the ODE solution $\state(t_i)$, which is known as Continuous Sensitivity Analysis and has a wide range of applications~\cite{rackauckas2018comparison}.

Historically, the two primary methods for computing derivatives through complex systems have been numerical and analytical derivation. Analytical (symbolic) derivation gives the user a chance to hand-optimize calculations, but is inflexible and error-prone, as gradients of any new dynamics elements must be determined separately.

\subsection{Numerical Differentiation}
A numerical approximation to the analytical gradients can be obtained through finite differences.
This approach is an one-at-a-time method (OAT) that, along each parameter dimension $d\in O(|\params|)$, adapts the parameter vector to approximate the gradient w.r.t. the final system state $\state(t_T)$. A common finite differencing approach is the symmetric difference quotient
$
\frac{g(y+h) - g(y-h)}{2h}
$
that approximates the gradient of function $g$ at point $y$ symmetrically at two nearby points to $y$ using the step size $h$. Its error is characterized as $O(h^2)$, while higher-order symmetric derivatives can be obtained that achieve higher accuracy at the cost of more function evaluations. In practice, step size $h$ cannot be reduced indefinitely due to floating point errors~\cite{jerrel1997ad}, limiting the overall achievable accuracy of this method.

\subsection{Automatic Differentiation}

Additionally, we may compute gradients by using forward or reverse mode automatic differentiation (AD) on the numerical integrator used to solve the ODE. Forward-mode differentiation performs arithmetic on dual numbers to compute functions and their derivatives simultaneously. Reverse-mode differentiation tracks the derivatives of function evaluations, storing them on a tape. Gradients are then computed in reverse by repeatedly applying the chain rule. Tape-based AD software may see high memory usage, especially for solvers with adaptive step sizes or systems with many outputs, while forward-mode differentiation scales poorly with input parameters. We present results for automatically differentiating ODE integrators in Section~\ref{sec:experiments}.

\subsection{Local Sensitivity Analysis}

Local sensitivity analysis, is a method for computing gradients for ODE solutions by augmenting the model dynamics to include the dynamics for the gradient itself (local sensitivities)
\begin{equation*}
    \frac{d}{dt}\left( \frac{\partial \state}{\partial \params_i} \right) =
    \frac{\partial f}{\partial \state}
    \frac{\partial \state}{\partial \params_i} +
    \frac{\partial f}{\partial \params_i}
\end{equation*}
where $\params_i$ is the $i$-th parameter.
This approach adds a new equation to the system per parameter, and thus performs poorly for systems with many parameters. Fortunately, for many applications in robotics, we are interested in optimizing a few unknown parameters of a model.

\begin{algorithm}
\begin{algorithmic}
\Require
  \Statex Mechanism parameters $\params$
  \Statex Initial state $\state(t_0)$
  \Statex Start time $t_0$, end time $t_1$
\Statex
\State $s_0 \gets [\state(t_0), \mathbf{0}_{|\state| \times |\params|}]$
\Function{augment}{$[\state(t), \pdv{\state(t)}{\params}], t, \params$}
    \State $A \gets \pdv{f}{\state(t)}\pdv{\state(t)}{\params} + \pdv{f}{\params}$
    \State\Return $[f(\state(t), t, \params), [A_{11}, \dots, A_{|\state||\params|}]]$
\EndFunction
\State $[\state(t_1), \pdv{\state(t_1)}{\params}]$
       $\gets\Call{Integrate}{s_0, \textsc{augment}, t_0, t_1, \params}$
\State\Return $\pdv{\state(t_1)}{\params}$
\end{algorithmic}
\label{algo:coupled}
\caption{Coupled ODE System}
\end{algorithm}

\subsection{Adjoint Sensitivity Analysis}

\newcommand{\adj}{\mathbf{a}}

To compute derivatives in our simulator, we use techniques from automatic differentiation and sensitivity analysis. There are multiple ways to compute gradients of functions of the solutions to a system of differential equations. A concise overview is given by \cite{rackauckas2018comparison}. One method, introduced by \cite{pontryagin1962mathematical} and popularized recently by \cite{chen2018neural}, is a continuous method called Adjoint Sensitivity Analysis.

We can compute a parameter gradient by backwards solving
\begin{equation*}
    \frac{d \adj(t)}{dt} = -\adj(t)^\intercal \frac{\partial f(\state(t), t, \params)}{\partial \state},
\end{equation*}
which is known as the adjoint problem.
At every discrete point $t_i$ where cost is evaluated, the ODE solution is perturbed by $\frac{\partial \Loss}{\partial \state(t)}$ where $\state(t)$ is solved in the forward pass. Then the loss gradient is\todo{verify}
\begin{align*}
\frac{d\Loss}{d\params} &= \adj(t_0)^\intercal\frac{\partial f(\state(t_0), \params, t_0)}{\partial \state} \\
&\ +\sum_i  \int_{t_i}^{t_{i+1}}  \adj(t)^\intercal\frac{\partial f(\state(t), \params, t)}{\partial \params} dt
\end{align*}

\begin{algorithm}
\begin{algorithmic}
\Require
  \Statex Mechanism parameters $\params$
  \Statex Final state $\state(t_1)$
  \Statex Loss gradient of the final state $\adj(t_1) = \pdv{\Loss}{\state(t_1)}$
  \Statex Start time $t_0$, end time $t_1$
\Statex
\State $s_0 \gets [\state(t_1), \pdv{\Loss}{\state(t_1)}, \mathbf{0}_{|\params|}]$
\Function{augment}{$[\state(t), \adj(t), \cdot], t, \params$}
    \State\Return $[f(\state(t), t, \params), -\adj(t)^\intercal\pdv{f}{\state}, -\adj(t)^\intercal\pdv{f}{\params}]$
\EndFunction
\State $[\state(t_0), \pdv{\Loss}{\state(t_0)}, \pdv{\Loss}{\params}]$
       $\gets\Call{Integrate}{s_0, \textsc{augment}, t_1, t_0, \params}$
\State\Return $\pdv{\Loss}{\params}$
\end{algorithmic}
\label{algo:asm}
\caption{Adjoint Sensitivity Method}
\end{algorithm}

\section{EXPERIMENTS}
\label{sec:experiments}
We evaluate the previously introduced methods for computing gradients through the ODE solutions. To this end, we first benchmark these approaches and subsequently compare them on parameter estimation problems involving a simulated compound pendulum and the automatic design of a robot arm. Next, we present an adaptive control algorithm that leverages the parameter estimation capabilities of our differentiable simulator and combines it with trajectory optimization to control a mechanism in a different simulator. Our simulator is implemented in C++ using the Eigen framework~\cite{eigenweb} for linear algebra and Stan Math~\cite{carpenter2015stan} for automatic differentiation (AD). Since the latter only provides an implementation of reverse-mode automatic differentiation, we limit our attention to this algorithm and leave considerations of forward-mode and other AD techniques for future work.

\subsection{Benchmarking Gradient Calculation Approaches}
In our first experiment, we consider an $n$-link compound pendulum that is simulated over a variety of time steps given its link lengths $l_k$ as parameters $\params=\{l_1,...,l_n\}$. We focus in our profiling of the approaches introduced in \autoref{sec:approach} on their computational efficiency, i.e., how many dynamics evaluations $f(\cdot)$ are necessary, how many variables are generated on the automatic differentiation stack, and the total computation time.

\begin{figure*}
    \centering
    \newcommand{\benchwidth}{0.24\textwidth}
    \newcommand{\supersize}{0.7cm 0 0.7cm 0}
    \includegraphics[width=\benchwidth,trim=0.7cm 0 0.7cm 0]{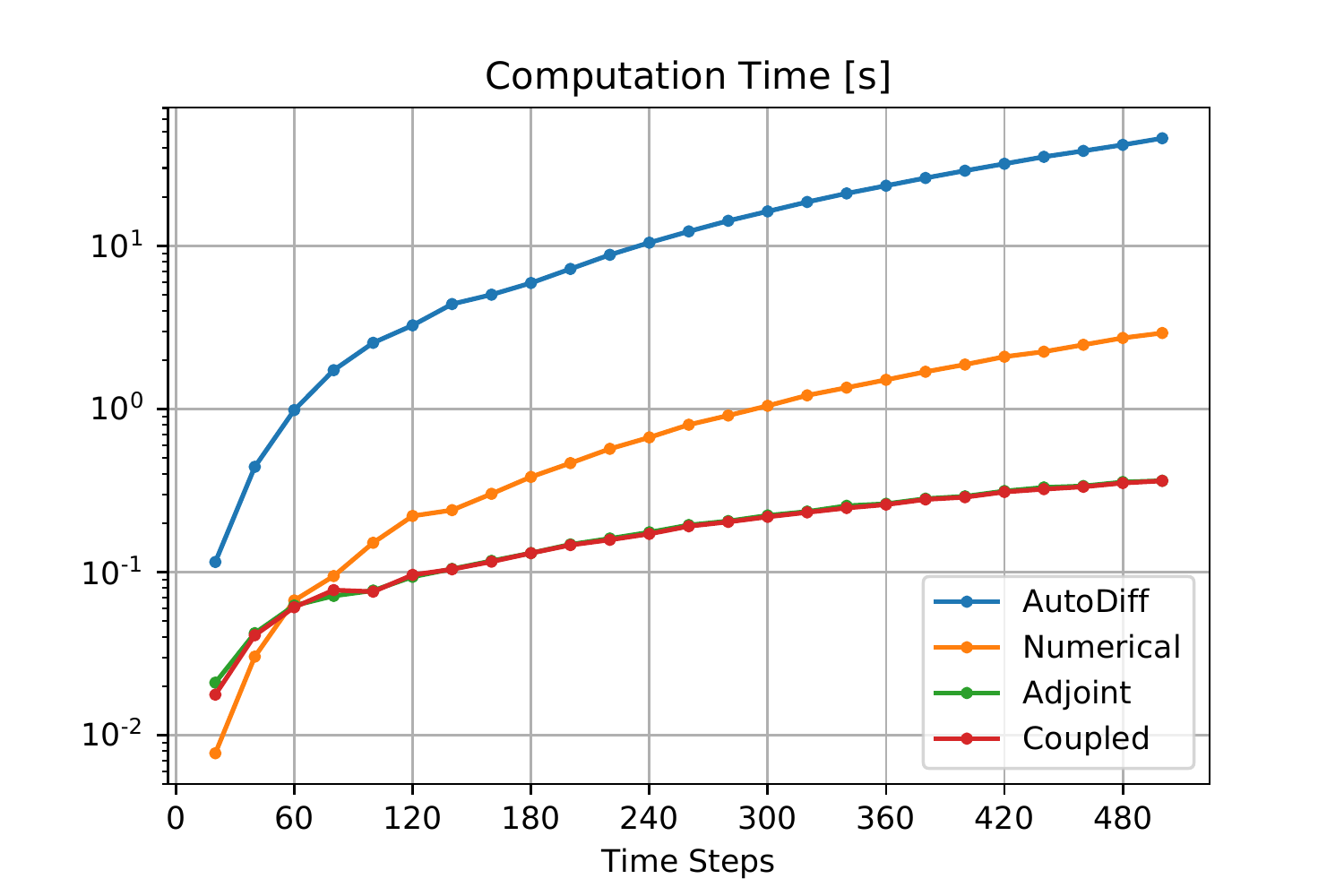}
    \includegraphics[width=\benchwidth,trim=0.7cm 0 0.7cm 0]{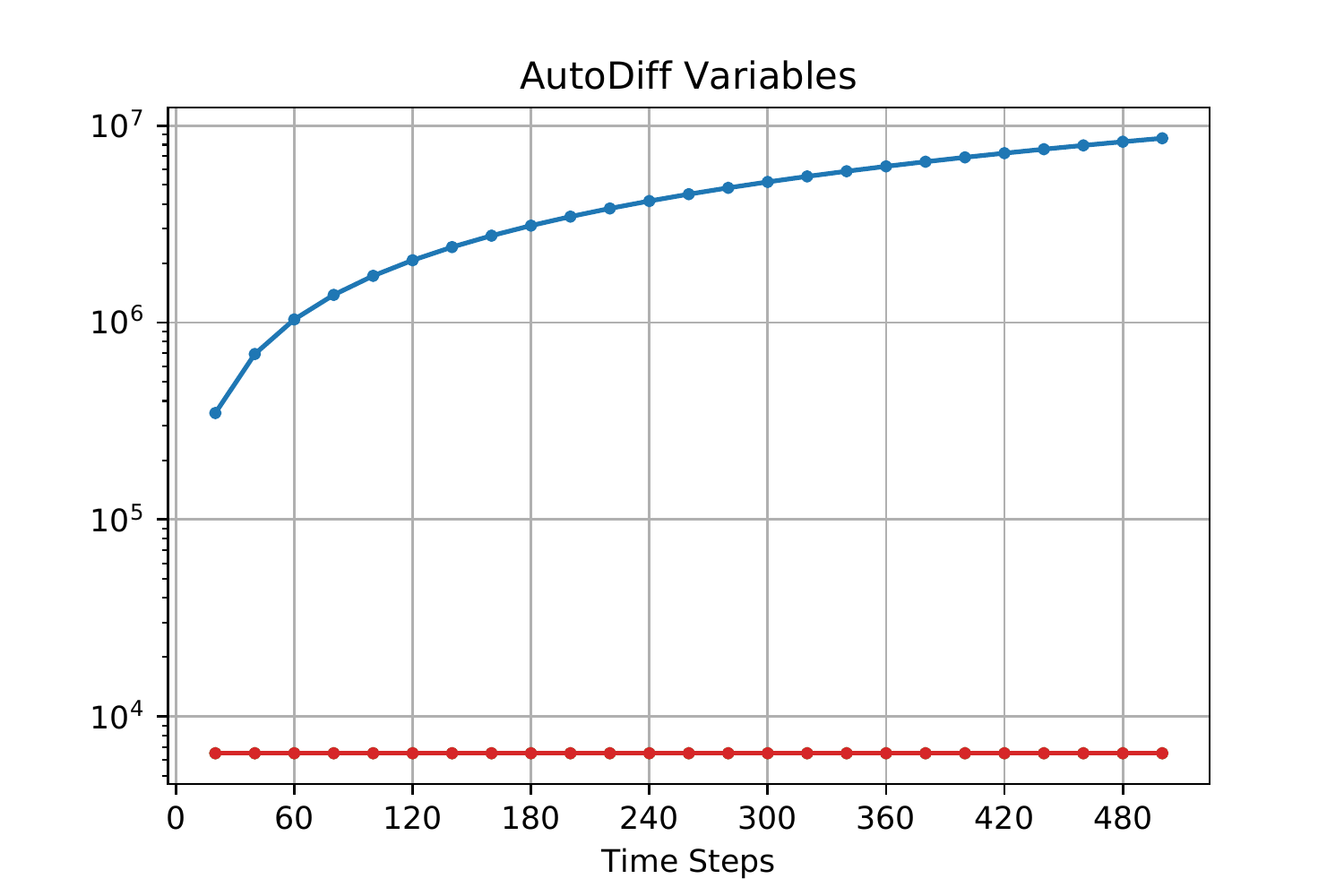}
    \includegraphics[width=\benchwidth,trim=0.7cm 0 0.7cm 0]{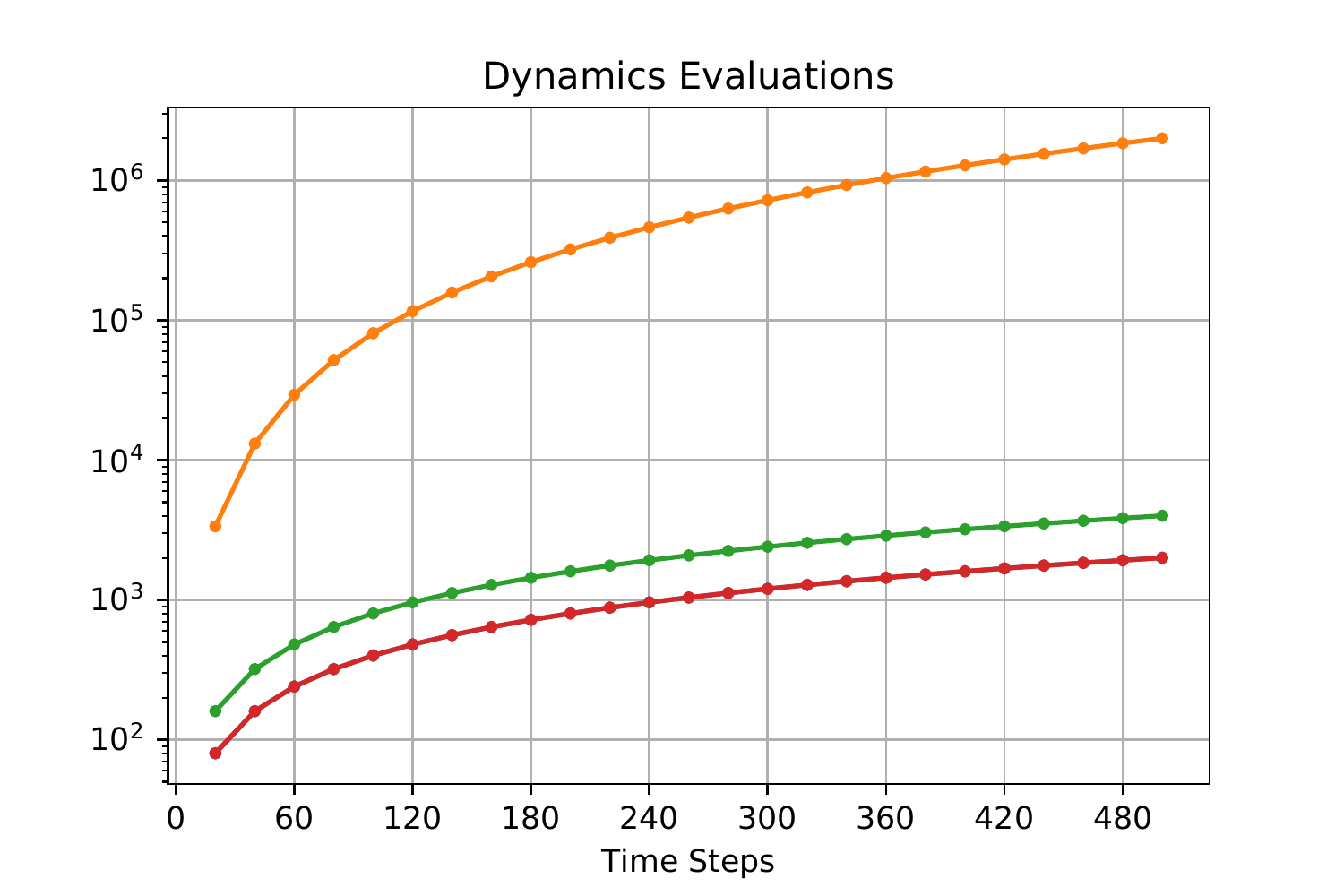}\hfill
    \includegraphics[width=\benchwidth,trim=0.7cm 0 0.7cm 0]{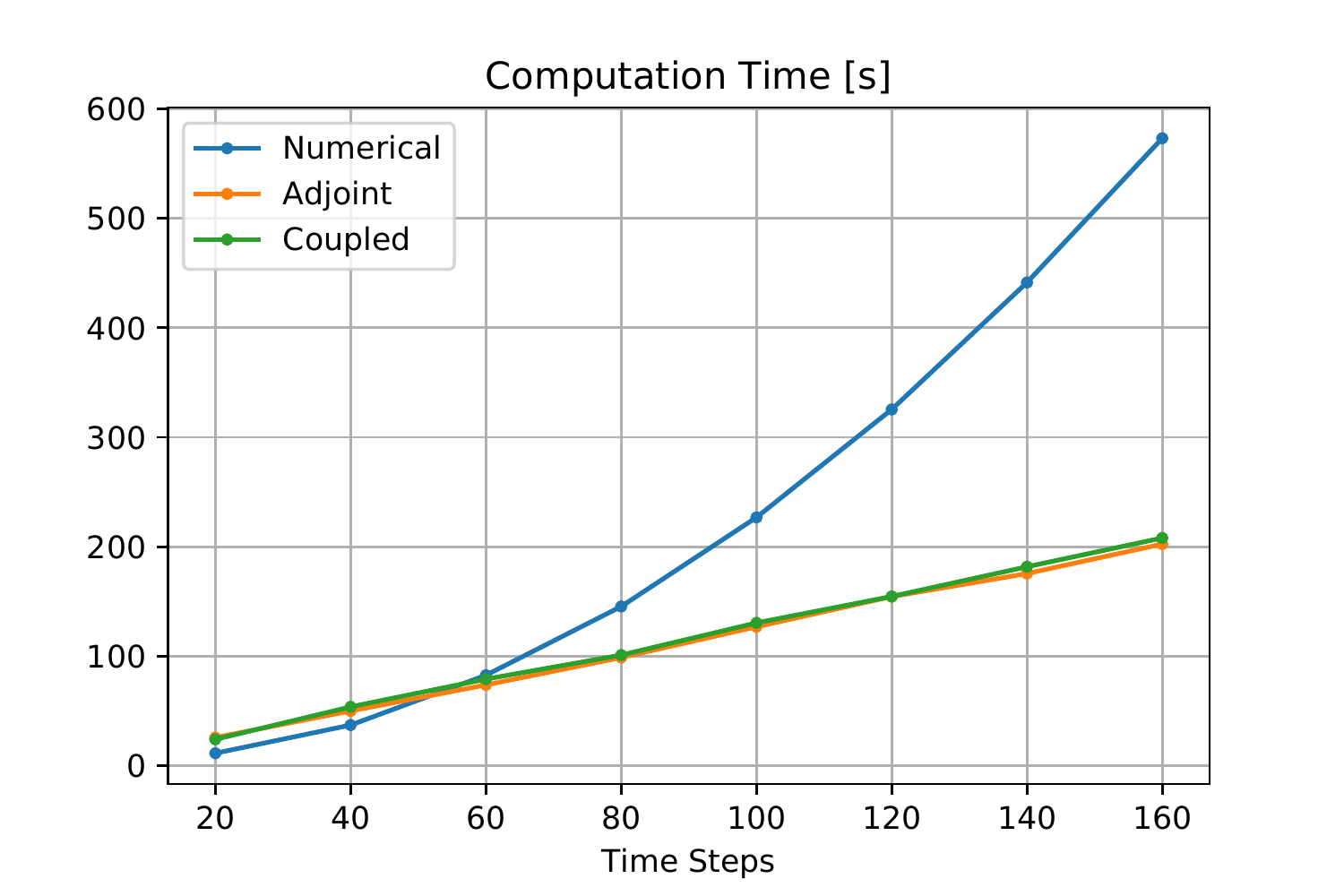}
    \caption{Benchmarking results of the four Jacobian calculation methods considered in this work on the fourth-order Runge-Kutta integrator. \textit{First three plots:} results for a double pendulum (logarithmic scale). \textit{Last plot:} computation times for a 100-link compound pendulum.}
    \label{fig:benchmarking_rk4}
\end{figure*}

We first consider a double pendulum ($n=2$) and report the performance of the algorithms Numerical Differentiation (``Numerical''), reverse-mode AD (``AutoDiff''), Adjoint Sensitivity Method (``Adjoint'') and Coupled ODE System (``Coupled'') in \autoref{fig:benchmarking_rk4}. Although the number of ODE evaluations (third plot) grows exponentially faster with finite differencing (orange) than the other methods, we note that AutoDiff takes the longest. We conduct the experiment using error-controlled adaptive time stepping methods, such as the Dormand-Price and the Fehlberg methods, and observe the same behavior. Reverse-mode AD records a copy of each participating variable \emph{per operation}, resulting in a large stack of variables (second plot), that needs to be traversed in order to compute gradients. In contrast, Adjoint and Coupled both maintain a constant-size stack of variables while taking approximately the same computation time, while the later requires approximately twice as many dynamics evaluations as the former. Next, we investigate the scalability of the continuous sensitivity methods on a 100-link compound pendulum, requiring a parameter vector of size 100 to be estimated. We exclude AD from this comparison due to its prohibitively high computation time. Adjoint and Coupled both remain close in computation time, although Coupled requires $|\state|\times|\theta|$ sensitivites (\autoref{algo:coupled}) compared to Adjoint's $|\theta|$ augmented state dimensions (\autoref{algo:asm}), which might be offset due to the need of solving two ODE in the case of Adjoint at each time step.

\subsection{Automatic Robot Design}
Industrial robotic applications often require a robot to follow a given path in task space with its end effector. In general, robotic arms with six or more degrees of freedom provide large workspaces and redundant configurations to reach any possible point within the workspace. However, motors are expensive to produce, maintain, and calibrate. Designing arms that contain a minimal number of motors required for a task provides economic and reliability benefits, but imposes constraints on the kinematic design of the arm.

One standard for specifying the kinematic configuration of a serial robot arm is the Denavit-Hartenberg (DH) parameterization \cite{hartenberg1955kinematic}. For each joint $i$, the DH parameters are $(d_i, \theta_i, a_i, \alpha_i)$, describing the distance from joint $i$ to motor axis $i-1$, the rotation about axis $i-1$, the distance of joint $i$ along motor axis $i-1$, and the angle between motor axes $i$ and $i-1$, respectively.

\begin{figure}[!b]
    \centering
    \includegraphics[width=\columnwidth]{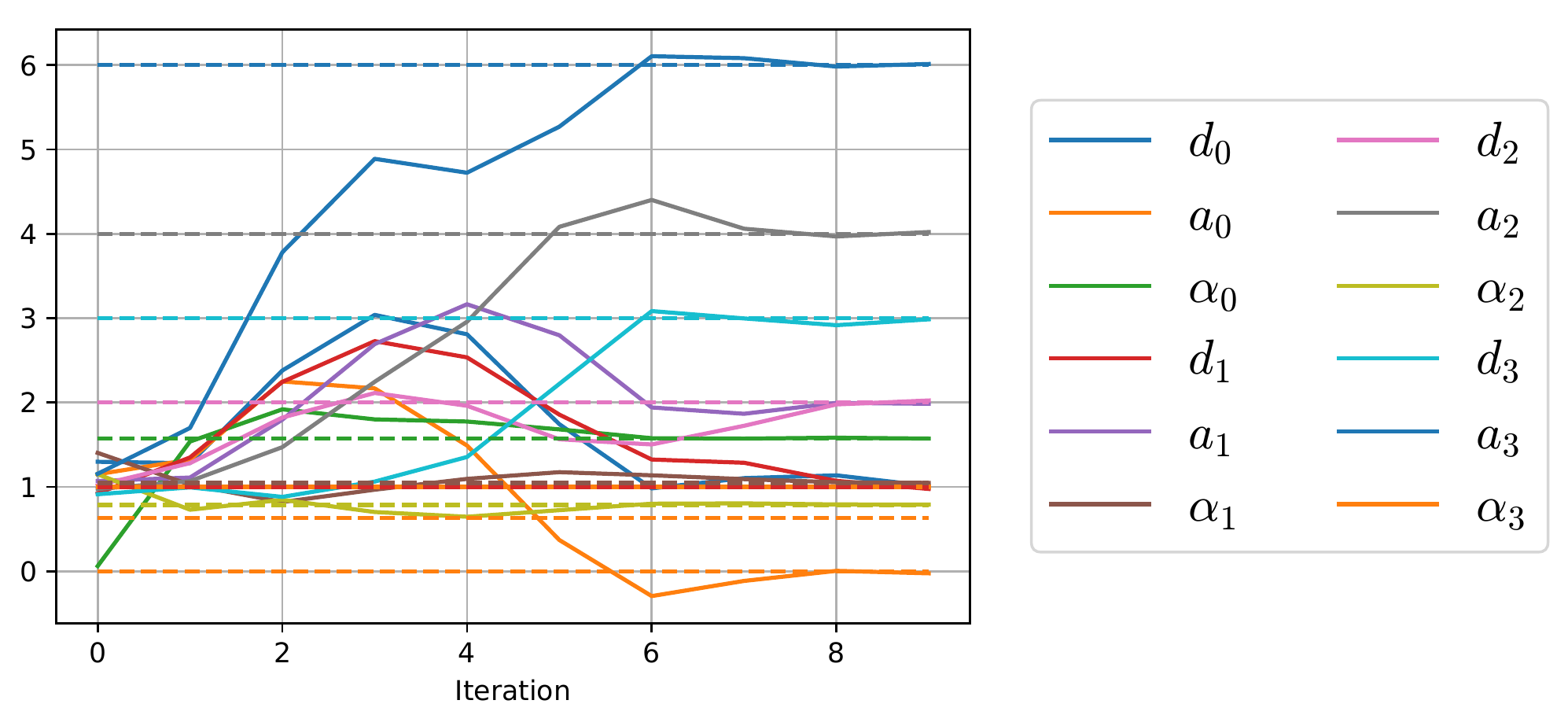}
    \caption{Optimization of a 4-DOF robot arm design parameterized by the Denavit-Hartenberg (DH) parameters to match the robot arm that generated a given trajectory. (left) Evolution of the DH parameters over the optimization iterations.}
    \label{fig:dh_params}
\end{figure}

We specify a task-space trajectory $[ \mathbf{p}_0, \mathbf{p}_1, \dots, \mathbf{p}_T ]$ for $\mathbf{p}_t \in \mathbb{R}^3$ as the position in world coordinates of the robot's end-effector. Given a joint-space trajectory $[ \Q_0, \Q_1, \dots, \Q_T ]$, we seek to find the best $N$-DOF robot arm design, parameterized by DH parameters $\params^*\in\mathbb{R}^{3N}$, that most closely matches the specified end-effector trajectory:
\begin{align*}
    \params^* = \argmin_{\params} \sum_{t=0}^{T} ||\operatorname{KIN}_{\params}(\Q_t) - \mathbf{p}_t||_2^2,
\end{align*}
where the forward kinematics function $\operatorname{KIN}(\cdot)$ maps from joint space to the Cartesian coordinates of the end-effector, conditioned on DH parameters $\params$. Since we compute $\operatorname{KIN}(\cdot)$ using our engine, we may compute derivatives of arbitrary inputs to this function, and use gradient-based optimization through L-BFGS~\cite{liu1989lbfgs} from the Ceres optimization library~\cite{ceres-solver} to converge to arm designs which accurately perform the trajectory-following task, as shown in \autoref{fig:dh_params}.

\begin{figure}[!b]
    \centering
    \includegraphics[width=.5\columnwidth,trim=22cm 1cm 14cm 3cm,clip]{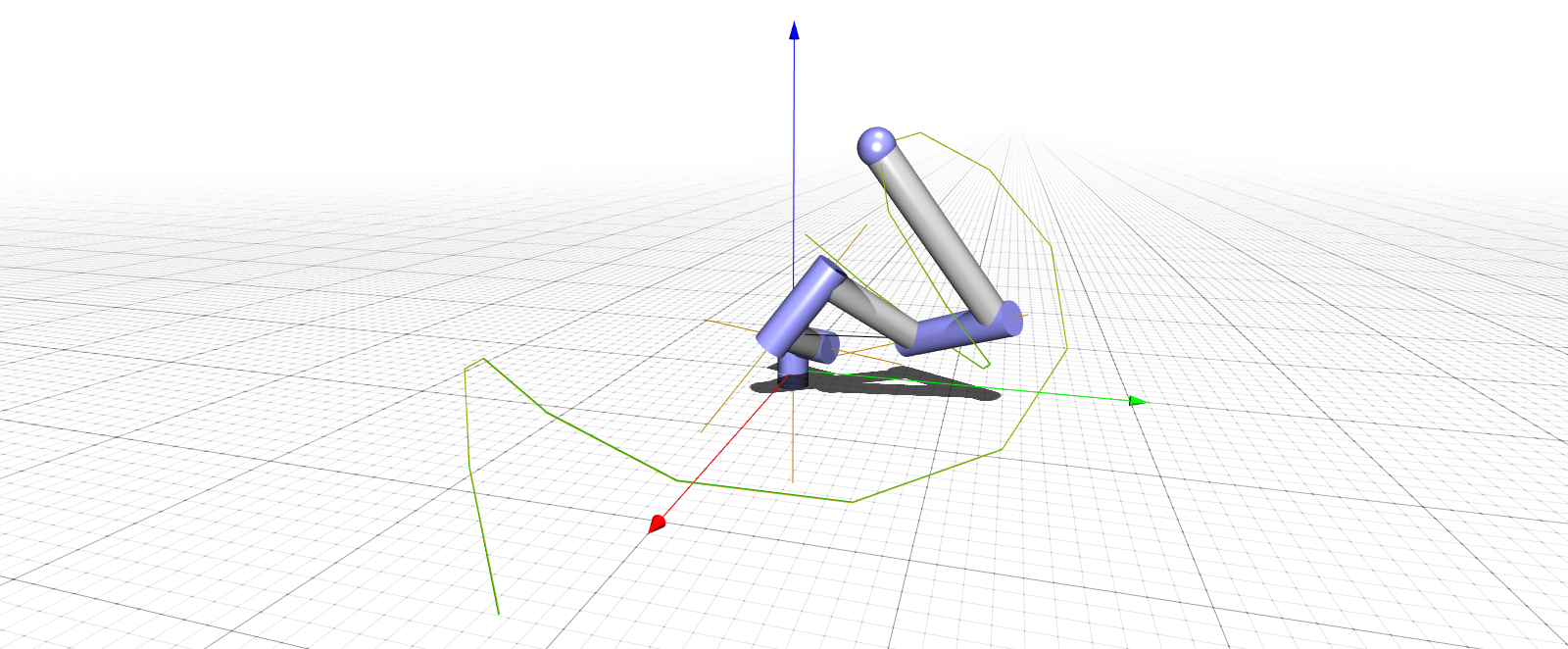}
    \caption{Visualization of an exemplary 4-DOF robot arm and its trajectory in our simulator.}
    \label{fig:dh_example}
\end{figure}

\begin{algorithm*}
\begin{algorithmic}
\Require Cost function $\boldsymbol{J}$, episode length $M$, trajectory length $T$, horizon length $H$
\For{episode = $1..M$}
    \State $R\gets \emptyset$
    \qquad\Comment{Replay buffer to store transition samples from the real environment}
    \State Obtain initial state $\state^*_{0}$ from the real environment
    \For{$t=1..T$}
        \State $\displaystyle \{\control^*\}_{t}^{t+H} \gets \argmin_{\control_{1:H}} \boldsymbol{J}$
        \Comment{Trajectory optimization using iLQR with cost from \autoref{eq:ampc_cost}}
        \State \qquad\qquad\ \ \ \ s.t. $\state_1 = \state^*_{t},\ \ \state_{i+1}=\int f([\state_i, \control_i], i, \params),\ \ \underbar{\control}\leq \control \leq \bar{\control}$
        \State Take action $\control^*_t$ in the real environment and obtain next state $\state^*_{t+1}$
        \State Store transition $(\state^*_t, \control^*_t, \state^*_{t+1})$ in $R$
    \EndFor
    \State Fit dynamics model $f$ to real data $R$ by minimizing the state-action prediction loss (\autoref{eq:xu_loss})
\EndFor
\end{algorithmic}
\caption{Adaptive MPC algorithm using differentiable physics model $f_{\params}$.}
\label{algo:ampc}
\end{algorithm*}

\subsection{Adaptive MPC}
\label{sec:ampc}
Besides parameter estimation and design, a key benefit of differentiable physics is its applicability to optimal control algorithms. In order to control a system within our simulator, we specify the control space $\ControlSpace$, which is typically a subset of the system's generalized forces $\forces$, and the state space $\StateSpace$. Given a quadratic, i.e. twice-differentiable, cost function \mbox{$c:\ \StateSpace\times\ControlSpace\to\mathbb{R}$}, we can linearize the dynamics at every time step, allowing efficient gradient-based optimal control techniques to be employed. Iterative Linear Quadratic Control~\cite{li2004iterative} (iLQR) is a direct trajectory optimization algorithm that uses a dynamic programming scheme on the linearized dynamics to derive the control inputs that successively move the trajectory of states and controls closer to the optimum of the cost function.

Throughout our control experiments, we optimize a trajectory for an $n$-link cartpole to swing up from an arbitrary initial configuration of the joint angles. In the case of double cartpole, i.e. a double inverted pendulum on a cart, the state $\state\in\StateSpace$ is defined as
$$
\state = \left(p, \dot{p}, \sin q_0, \cos q_0, \sin q_1, \cos q_1, \dot{q}_0, \dot{q}_1, \ddot{q}_0, \ddot{q}_1\right),
$$
where $p$ and $\dot{p}$ refer to the cart's position and velocity, $(q_0, q_1) = \Q$ to the joint angles, and $(\dot{q}_0, \dot{q}_1)=\Qd, (\ddot{q}_0, \ddot{q}_1)=\Qdd$ to the velocities and accelerations of the revolute joints of the poles, respectively. For a single cartpole the state space is represented analogously, excluding the second revolute joint coordinates $q_1,\dot{q}_1,\ddot{q}_1$. The control input $\control\in\ControlSpace$ is a one-dimensional vector describing the force applied to the cart along the $x$ axis. As typical for finite-horizon, discrete-time LQR problems, the cost of a trajectory over $H$ time steps is defined as
\begin{align}
    \label{eq:ampc_cost}
    \boldsymbol{J} = \sum_{k=0}^{H-1}(\tilde{\state}_k^\text{T} Q \tilde{\state}_k + \control_k^\text{T} R \control_k) + \tilde{\state}_H^\text{T} S \tilde{\state}_H,
\end{align}
where $\tilde{\state}_k = \state^*-\state_k$, and the matrices $Q, S \in\mathbb{R}^{|\state|\times|\state|}$ and $R \in\mathbb{R}^{|\control|\times|\control|}$ weight the contributions of each dimension of the state and control input. Throughout this experiment, we set $Q, S, R$ to be diagonal matrices. Minimizing the cost function drives the system to the defined goal state\footnote{The goal state is given for a double cartpole here, it is analogously defined for a single cartpole.}
$
\state^* = \left(0, 0, 0, 1, 0, 1, 0, 0, 0, 0\right),
$
at which the pole is upright at zero angular velocity and acceleration, and the cart is centered at the origin with zero positional velocity.

Trajectory optimization assumes that the dynamics model is accurate w.r.t the real world and generates sequences of actions that achieve optimal behavior toward a given goal state, leading to open-loop control. Model-predictive control (MPC) leverages trajectory optimization in a feedback loop where the next action is chosen as the first control computed by trajectory optimization over a shorter time horizon with the internal dynamics model. After some actions are executed in the real world and subsequent state samples are observed, \emph{adaptive} MPC (Algorithm~\ref{algo:ampc}) fits the dynamics model to these samples to align it closer with the real-world dynamics. In this experiment, we want to investigate how differentiable physics can help overcome the domain shift that poses an essential challenge of model-based control algorithms that are employed in a different environment. To this end, we incorporate our simulator as dynamics model in such receding-horizon control algorithm to achieve swing-up motions of a single and double cartpole in the DeepMind Control Suite~\cite{tassa2018dm} environments that are based on the MuJoCo physics simulator.

\begin{figure}
    \centering
    \includegraphics[width=\columnwidth]{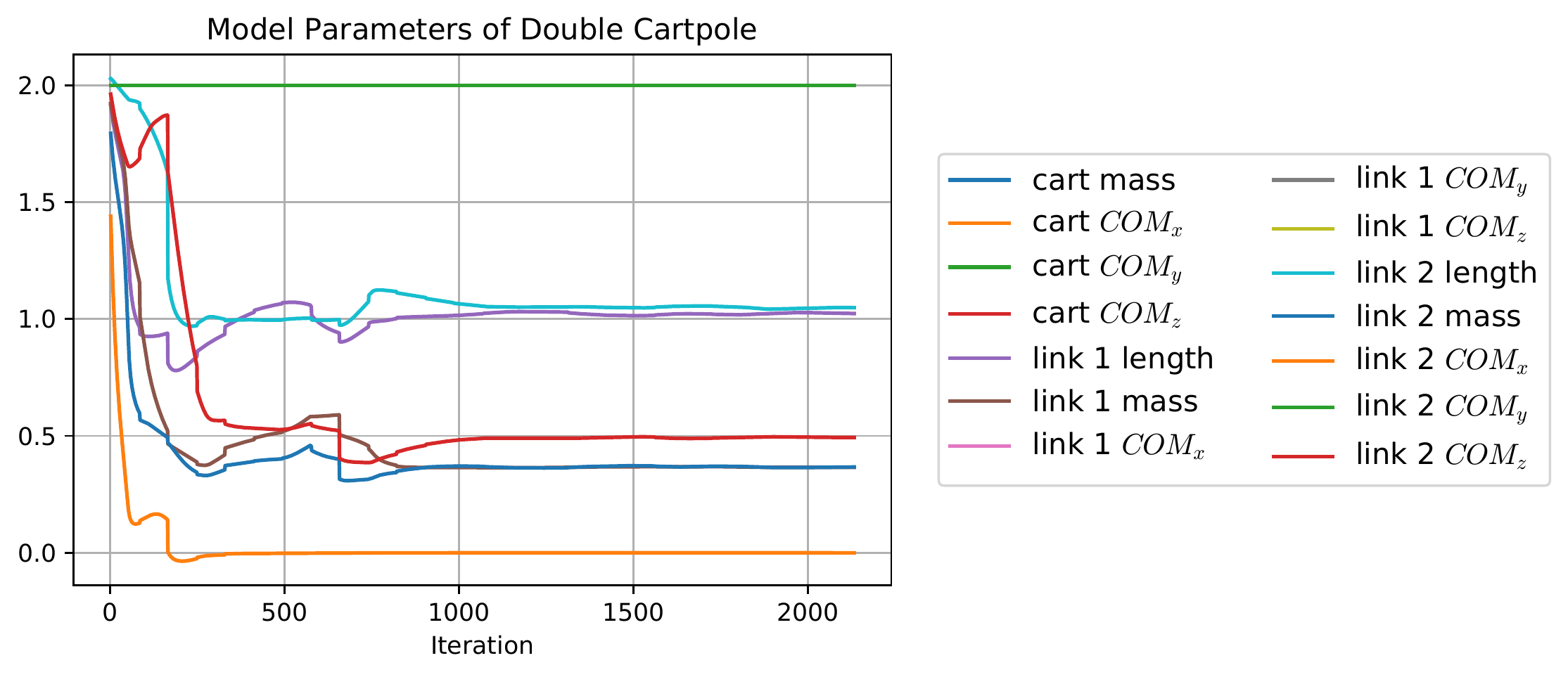}\vspace{-2em}
    \caption{Convergence of the physical parameters of a double cartpole, over all model fitting iterations combined, using Adaptive MPC (Algorithm~\ref{algo:ampc}) in the DeepMind Control Suite environment.}
    \label{fig:ampc_cartpole_param}
\end{figure}

We fit the parameters $\params$ of the simulator by minimizing the prediction loss given the state-action transition $(\state^*_t, \control^*_t, \state^*_{t+1})$ from the real system:
\begin{equation}
    \params^* = \argmin_{\params} \sum_{t} ||\int f ([\state^*_t, \control^*_t], t, \params) - \state^*_{t+1}||_2^2
\label{eq:xu_loss}
\end{equation}
Thanks to the low dimensionality of the model parameter vector $\params$ (for a double cartpole there are 14 parameters, cf. \autoref{fig:ampc_cartpole_param}), efficient optimizers such as the quasi-Newton optimizer L-BFGS are applicable, leading to fast convergence of the fitting phase, typically within 10 optimization steps. The length $T$ of one episode is 140 time steps. During the first episode we fit the dynamics model more often, i.e. every 50 time steps, to warm-start the receding-horizon control scheme. Given a horizon size $H$ of 20 and 40 time steps, MPC is able to find the optimal swing-up trajectory for the single and double cartpole, respectively.

%\begin{figure*}
%    \centering
%    \includegraphics[width=.48\textwidth,trim=0 8cm 0 0,clip]{fig/fitting_InertiaCompoundCartpole1_ep00_fit00.pdf}\hfill
%    \includegraphics[width=.48\textwidth,trim=0 8cm 0 0,clip]{fig/fitting_InertiaCompoundCartpole1_ep02_fit00.pdf}
%    \vspace{-2em}
%    \caption{Trajectory of the six-dimensional cartpole state over 140 time steps after the first (left) and third (right) episode of Adaptive MPC (\autoref{algo:ampc}). After two episodes, the differentiable physics model (orange) has converged to model parameters that allow it to accurately predict the cartpole dynamics modelled in MuJoCo (blue). Since by the third episode the control algorithm has converged to a successful cartpole swing-up, the trajectory differ significantly from the first roll-out.}
%    \label{fig:rl_mpc_fitting}
%\end{figure*}

Within a handful of training episodes, adaptive MPC infers the correct model parameters involved in the dynamics of a double cartpole (\autoref{fig:ampc_cartpole_param}). As shown in \autoref{fig:models}, the models we start from do not match their counterparts from DeepMind Control Suite. For example, the poles are represented by capsules where the mass is distributed across these elongated geometries, whereas initially in our model, the center of mass of the links is at the end of them, such that they have different inertia parameters. We set the masses, lengths of the links, and 3D coordinates of the center of masses to 2, and, using a few steps of the optimizer and less than 100 transition samples, converge to a much more accurate model of the true dynamics in the MuJoCo environment.
% On the example of a cartpole, \autoref{fig:rl_mpc_fitting} visualizes the predicted and actual dynamics for each state dimension after the first (left) and third (right) episode.

\section{CONCLUSION}
\label{sec:conclusion}

We introduced a novel differentiable physical simulator, and presented experiments for the inference of physical parameters, optimal control and system design. Since it is constrained to the laws of physics, such as conservation of energy and momentum, our proposed model provides a large, meaningful inductive bias on robot learning problems. 
Within a handful of trials in out test environment, our gradient-based representation of rigid-body dynamics allows an adaptive MPC scheme to infer the model parameters of the system thereby allowing it to make predictions and plan for actions many time steps ahead.
We look forward to exercising this physics engine for learning and control to solve complex tasks on physical robot systems.

% \section*{ACKNOWLEDGMENT}

% The preferred spelling of the word ÒacknowledgmentÓ in America is without an ÒeÓ after the ÒgÓ. Avoid the stilted expression, ÒOne of us (R. B. G.) thanks . . .Ó  Instead, try ÒR. B. G. thanksÓ. Put sponsor acknowledgments in the unnumbered footnote on the first page.

% %%%%%%%%%%%%%%%%%%%%%%%%%%%%%%%%%%%%%%%%%%%%%%%%%%%%%%%%%%%%%%%%%%%%%%%%%%%%%%%%

% References are important to the reader; therefore, each citation must be complete and correct. If at all possible, references should be commonly available publications.

\newpage

\bibliographystyle{IEEEtran}
\bibliography{IEEEabrv,literature}

\end{document}